\documentclass[11pt,a4paper]{article}
\usepackage[hyperref]{acl2021}
\usepackage{times}
\usepackage{latexsym}
\usepackage{url}

\usepackage[]{graphicx}
\usepackage{caption}
\usepackage{subcaption}

\usepackage{microtype}

\aclfinalcopy % Uncomment this line for the final submission
\def\aclpaperid{2766} %  Enter the acl Paper ID here

\usepackage{color, colortbl}
\definecolor{LightRed}{rgb}{0.88,1,1}

\usepackage{xspace,mfirstuc,tabulary}
\newcommand{\glove}{\textsc{GloVe}\xspace}
\newcommand{\cbow}{\textsc{CBOW}\xspace}
\newcommand{\skipgram}{\textsc{Skipgram}\xspace}
\newcommand{\stov}{\textsc{Sent2vec}\xspace}

\newcommand{\wtov}{\textsc{Word2Vec}\xspace}

\newcommand{\bert}{\textsc{BERT}\xspace}
\newcommand{\roberta}{\textsc{RoBERTa}\xspace}
\newcommand{\gpttwo}{\textsc{GPT2}\xspace}

\newcommand{\bertbase}{\textsc{BERT}-\textsc{12}\xspace}
\newcommand{\sbertbasenli}{\textsc{sBERT}-\textsc{base}-\textsc{NLI}\xspace}
\newcommand{\bertlarge}{\textsc{BERT}-\textsc{24}\xspace}
\newcommand{\robertabase}{\textsc{RoBERTa}-\textsc{12}\xspace}
\newcommand{\srobertabasenli}{\textsc{sRoBERTa}-\textsc{base}-\textsc{NLI}\xspace}

\newcommand{\robertalarge}{\textsc{RoBERTa}-\textsc{24}\xspace}

\newcommand{\gpttwosmall}{\textsc{GPT2}-\textsc{12}\xspace}
\newcommand{\gpttwomedium}{\textsc{GPT2}-\textsc{24}\xspace}

\newcommand{\xtostatic}{\textsc{X2Static}\xspace}
\newcommand{\berttostatic}{\textsc{BERT2Static}\xspace}
\newcommand{\robertatostatic}{\textsc{RoBERTa2Static}\xspace}
\newcommand{\gpttwotostatic}{\textsc{GPT$_2$2Static}\xspace}

\newcommand{\fasttext}{\textsc{FastText}\xspace}

\newcommand{\ASEblank}{\textsc{ASE}\xspace}

\usepackage{todonotes}
\usepackage{booktabs}
\usepackage{multirow}

\title{Obtaining Better Static Word Embeddings \\
Using Contextual Embedding Models}

\author{Prakhar Gupta \\
  EPFL, Switzerland\\
  \texttt{prakhar.gupta@epfl.ch} \\\And
  Martin Jaggi \\
  EPFL, Switzerland \\
  \texttt{martin.jaggi@epfl.ch} \\}

\date{}

\usepackage{amsmath,amsfonts,bm}

\def\1{\bm{1}}

\def\vu{{\bm{u}}}
\def\vv{{\bm{v}}}

\def\mU{{\bm{U}}}
\def\mV{{\bm{V}}}

\DeclareMathAlphabet{\mathsfit}{\encodingdefault}{\sfdefault}{m}{sl}
\SetMathAlphabet{\mathsfit}{bold}{\encodingdefault}{\sfdefault}{bx}{n}

\def\cC{{\mathcal{C}}}

\begin{document}
\maketitle
\begin{abstract}
The advent of contextual word embeddings---representations of words which incorporate semantic and syntactic information from their context---has led to tremendous improvements on a wide variety of NLP tasks.
However, recent contextual models have prohibitively high computational cost in many use-cases and are often hard to interpret.
In this work, we demonstrate that our proposed distillation method, which is a simple extension of CBOW-based training,
allows to significantly improve computational efficiency of NLP applications,
while outperforming the quality of existing static embeddings trained from scratch as well as those distilled from previously proposed methods.
As a side-effect, our approach also allows a fair comparison of both contextual and static embeddings via standard lexical evaluation tasks.
\end{abstract}

\section{Introduction}
Word embeddings---representations of words which reflect semantic and syntactic information carried by them are ubiquitous in Natural Language Processing.
Static word representation models such as \glove \citep{Pennington2014GloveGV}, \cbow, \skipgram \citep{mikolov2013efficient} and \stov \citep{Pagliardini2018}
obtain stand-alone representations which do not depend on their surrounding words or sentences (context).
Contextual embedding models \citep{Devlin2019BERTPO, Peters2018DeepCW, Liu2019RoBERTaAR, Radford2019LanguageMA, schwenk-douze-2017-learning} on the other hand, embed the contextual information as well into the word representations making them more expressive than static word representations in most use-cases.

While recent progress on contextual embeddings has been tremendously impactful,
static embeddings still remain fundamentally important in many scenarios as well:
\begin{itemize}
   \item Even when ignoring the training phase, the computational cost of using static word embeddings is typically tens of millions times
   lower than using standard contextual embedding models\footnote{BERT base \cite{Devlin2019BERTPO} produces 768 dimensional word embeddings using 109M parameters,  requiring 29B FLOPs per inference call \cite{clark2020electra}.}%
, which is particularly important for latency-critical applications and on low-resource devices, and in view of environmental costs of NLP models 
\citep{Strubell2019EnergyAP}.
  \item Many NLP tasks inherently rely on static word embeddings \citep{shoemark-etal-2019-room}, for example for interpretability, or e.g. in research in bias detection and removal \citep{kaneko-bollegala-2019-gender,Gonen2019LipstickOA,Manzini2019BlackIT} and analyzing word vector spaces \citep{Vulic2020AreAG} or other metrics which are non-contextual by choice.
\item Static word embeddings can complement contextual word embeddings, for separating static from contextual semantics \citep{Barsalou1982ContextindependentAC,RubioFernndez2008ConceptNT}, or for improving joint embedding performance on downstream tasks \citep{alghanmi2020combining}.
\end{itemize}
We also refer the reader to this article\footnote{Do humanists need BERT? (\url{https://tedunderwood.com/2019/07/15/}) } illustrating several down-sides of using BERT-like models over static embedding models for non-specialist users. 
Indeed, we can see continued prevalence of static word embeddings in industry and research areas including but not limited to medicine \citep{Zhang2019BioWordVecIB,Karadeniz2019LinkingET,Magna2020ApplicationOM} and social sciences \citep{Rheault2020WordEF,GordonPolibias,Farrell2020OnTU,Lucy2020ContentAO}.

From a cognitive science point of view, Human language has been hypothesized to have both contextual as well as context-independent properties \citep{Barsalou1982ContextindependentAC,RubioFernndez2008ConceptNT} underlining the need for continued research in studying the expressiveness context-independent embeddings on the level of words. 

Most existing word embedding models, whether static or contextual, follow \citet{Firth1957ASO}'s famous hypothesis - ``You shall know a word by the company it keeps'' , i.e., the meaning of a word arises from its context. During training existing static word embedding models, representations of contexts are generally approximated using averaging or sum of the constituent word embeddings, which disregards the relative word ordering as well
as the interplay of information beyond simple pairs of words, thus losing most  contextual information. Ad-hoc remedies attempt to capture longer contextual information per word using higher order n-grams like bigrams or trigrams, and have been shown to improve the performance of static word embedding models \citep{Gupta2019BetterWE,Zhao2017Ngram2vecLI}. However, these methods are not scalable to cover longer contexts.  

In this work, we obtain improved static word embeddings by leveraging recent contextual embedding advances, namely by distilling existing contextual embeddings into static ones. Our proposed distillation procedure is inspired by existing \cbow-based static word embedding algorithms, but during training plugs in any existing contextual representation to serve as the context element of each word.

Our resulting embeddings outperform the current static embedding methods, as well as the current state-of-the-art static embedding distillation method on both unsupervised lexical similarity tasks as well as on downstream supervised tasks, by a significant margin. The resulting static embeddings remain compatible with the underlying contextual model used, and thus allow us to gauge the extent of lexical information carried by static vs contextual word embeddings.
We release our code and trained embeddings publicly on GitHub\footnote{\url{https://github.com/epfml/X2Static}}.

\section{Related Work}
A few methods for distilling static embeddings have already been proposed.
\citet{Ethayarajh2019HowCA} propose using contextual embeddings of the same word in a large number of different contexts. They take the first principal component of the matrix formed by using these embeddings as rows and use it as a static embedding. However, this method is not scalable in terms of memory (the embedding matrix scaling with the number of contexts) and computational cost (PCA).

\citet{Bommasani2020InterpretingPC} propose two different approaches to obtain static embeddings from contextual models.

\begin{enumerate}
  \item \textbf{Decontextualized Static Embeddings} - The word $w$ alone without any context, after tokenization into constituents $w_1,\ldots, w_n$
  is fed to the contextual embedding model denoted by $M$ and the resulting static embedding is given by $g(M(w_1),\ldots, M(w_n))$ where $g$ is
  a pooling operation. It is observed that these embeddings perform dismally on the standard static word embedding evaluation tasks.
  \item \textbf{Aggregated Static Embeddings} - Since contextual embedding models are not trained on a single word (without any context) as input,
  an alternative approach is to obtain the contextual embedding of the word $w$ in different contexts and then pool(max, min or average) the embeddings obtained from these different contexts. They observe that average pooling leads to the best performance. We refer to this method (with average pooling) as \emph{$\ASEblank$} throughout the rest of the paper. As we see in our experiments, the performance of \ASEblank embeddings saturates quickly with increasing size of the raw text corpus and is therefore not scalable.
\end{enumerate}

Other related work includes distillation of contextual word embeddings to obtain sentence embeddings \citep{reimers-gurevych-2019-sentence}. 
We also refer the reader to \citet{mickus2020you} for a discussion on the semantic properties of contextual models (primarily \bert) as well as \citet{rogers2020primer}, a survey on different works exploring the inner workings of \bert including its semantic properties.

\section{Proposed Method}
To distill existing contextual word representation models into static word embeddings, we augment a \cbow-inspired static word-embedding method as our 
anchor method to accommodate additional contextual information of the (contextual)
teacher model.
\stov \citep{Pagliardini2018} is a modification of the \cbow static word-embedding method which instead of a fixed-size context window uses the entire sentence to predict the masked word.
It also has the ability to learn n-gram representations along with unigram representations, allowing to better disentangle local contextual information from the static unigram embeddings.
\stov, originally meant to obtain sentence embeddings and later repurposed to obtain word representations \citep{Gupta2019BetterWE} was shown to outperform competing methods including
\glove \citep{Pennington2014GloveGV}, \cbow, \skipgram \citep{mikolov2013efficient} and \fasttext \citep{Bojanowski2016EnrichingWV} on word similarity evaluations.
For a raw text corpus~$\cC$ (collection of sentences), the training objective is given by
\begin{equation}
  \label{s2v:obj}
\min_{\mU,\mV} \ \sum_{S \in \cC}  \sum_{w_t \in S} f(\vu_{w_t}, E_{\text{\textsf{ctx}}}(S,w_t) )
\end{equation}
where $f(\vu,\vv):=\ell(\vu^\top \vv) +  \sum_{w' \in N} \ell(-\vu_{w'}^\top \vv)$.
Here, $w_t$ is the masked target word, $\mU$ and $\mV$ are
the target word embedding and the source 
 n-gram matrices respectively, $N$ is the set of negative target samples
and, $\ell: x \mapsto \log{(1 + e^{-x})}$  is the logistic loss function.

For \stov, the context encoder $E_{\text{\textsf{ctx}}}$ used in optimizing \eqref{s2v:obj} is simply given by the (static, non-contextual) sum of all vectors in the sentence without the target word,
\begin{equation}
  \label{s2v:ctx}
E_{\text{\textsf{ctx}}}(S,w_t) := \tfrac{1}{|R(S\setminus\{w_t\})|} \hspace{-3mm}\sum_{w \in R(S\setminus\{w_t\})}\hspace{-3mm} \vv_w \ ,
\end{equation}
where $R(S)$ denotes the optional expansion of the sentence $S$ from words to short n-grams, i.e., the context sentence embedding is obtained by averaging the embeddings of word n-grams in the sentence $S$.

\begin{table*}[!htb]
  \centering
  \resizebox{1.45\columnwidth}{!}{%
\begin{tabular}{c|c|c|c|c|c|c}
\hline
\begin{tabular}[c]{@{}c@{}}Epoch(s)\\ trained\end{tabular} & \begin{tabular}[c]{@{}c@{}}Max\\ Vocab.\\ Size\end{tabular} & \begin{tabular}[c]{@{}c@{}}Number\\ of \\ Negatives\\ Sampled\end{tabular} & \begin{tabular}[c]{@{}c@{}}Target Word\\ Subsampling\\ hyperparameter\end{tabular} & \begin{tabular}[c]{@{}c@{}}Minimum\\ Word Count\end{tabular} & \begin{tabular}[c]{@{}c@{}}Initial \\ Learning\\ Rate\end{tabular} & \begin{tabular}[c]{@{}c@{}}Batch\\ Size\end{tabular} \\ \hline
1                                                          & 750000                                                              & 10                                                                         & 5e-6                                                                            & 10                                                           & 0.001                                                              & 128                                                  \\ \hline
\end{tabular}}
\caption{{\textbf{Training hyperparameters used for training \xtostatic models}}}
\label{tab:xtostat_param}
\end{table*}

\begin{table*}[]
  \centering
  \resizebox{2\columnwidth}{!}{%
\setlength{\tabcolsep}{3pt}%default 6
\begin{tabular}{c|c|c|c|c|c|c|c|c|c}
\hline
Model    & \begin{tabular}[c]{@{}c@{}}Epoch(s)\\ trained\end{tabular} & \begin{tabular}[c]{@{}c@{}}Max\\ Vocab.\\ Size\end{tabular} & \begin{tabular}[c]{@{}c@{}}Number\\ of \\ Negatives\\ Sampled\end{tabular} & \begin{tabular}[c]{@{}c@{}}Target Word\\ Subsampling\\ hyperparameter\end{tabular} & \begin{tabular}[c]{@{}c@{}}Min.\\ Word \\ Count\end{tabular} & \begin{tabular}[c]{@{}c@{}}Initial \\ Learning\\ Rate\end{tabular} & \begin{tabular}[c]{@{}c@{}}Word \\  N-grams\end{tabular} & \begin{tabular}[c]{@{}l@{}}Character\\ N-grams\end{tabular} & \begin{tabular}[c]{@{}l@{}}Window\\ Size\end{tabular} \\ \hline
\stov & \{5,\textbf{10},15\}                                                & 750000                                                              & \{5,8,\textbf{10}\}                                                                 & $\{$1e-4, 5e-6, {1e-5}, \textbf{5e-6}$\}$                                                            & 10                                                           & 0.2                                                                & \{1,2,\textbf{3}\}                                                  & N.A.                                                        & N.A.                                                   \\ \hline
\skipgram & \{5,\textbf{10},15\}                                                & N.A.                                                                & \{\textbf{5},8,10\}                                                             & $\{$1e-4, 5e-6, \textbf{1e-5}, 5e-6$\}$                                                            & 10                                                           & 0.05                                                               & N.A.                                                  & \{N.A.,\textbf{3-6}\}                                                & \{2,5,\textbf{10}\}                                             \\ \hline
\cbow     & \{5,10,\textbf{15}\}                                                & N.A.                                                                & \{\textbf{5},8,10\}                                                             & $\{$1e-4, 5e-6, \textbf{1e-5}, 5e-6$\}$                                                            & 10                                                           & 0.05                                                               & N.A.                                                  & \{N.A.,\textbf{3-6}\}                                                & \{2,5,\textbf{10}\}                                             \\ \hline
\end{tabular}
}
\caption{{\textbf{Hyperparameter search space description for the training of \stov, \skipgram and \cbow models}: Best hyperparameters for the chosen model in our experiments are shown in bold. N.A. indicates not applicable.}}
\label{tab:baseline_param}
\end{table*}

We will now generalize the objective \eqref{s2v:obj} by allowing the use of arbitrary modern contextual representations $E_{\text{\textsf{ctx}}}$ instead of the static context representation as in \eqref{s2v:ctx}.
This key element will allow us to translate quality gains from improved contextual representations also to better static word embedding in the resulting matrix~$\mU$.
We propose two different approaches of doing so, which differ in the granularity of context used for obtaining the contextual embeddings.

\subsection{Approach 1 - Sentences as context}
Using contextual representations of all words in the sentence $S$ (or the sentence $S\setminus\{w_t\}$ without the target word) allows for a more refined representation of the context, and to take in account the word order as well as the interplay of information among the words of the context.

More formally, let $M(S,w)$ denote the output of a contextual embedding-encoder, e.g. \bert, corresponding to the word $w$ when a piece of text $S$ containing $w$ is fed to it as input. We let
$E_{\text{\textsf{ctx}}}(S,w)$ to be the average of all contextual embeddings of words~$w$ returned by the encoder,
\begin{equation}
  \label{B2W2V:sent}
 E_{\text{\textsf{ctx}}}(S,w_t) := \tfrac{1}{|S|}\sum_{w \in S} M(S,w)
\end{equation}
This allows for a more refined representation of the context as the previous representation did not take in account neither the word order nor the interplay of information among the words of the context.
Certainly, using $S_{m_{w_t}}$($S$ with $w_t$ masked) and $w$  would make for an even better word-context pair but that would amount to
one contextual embedding-encoder inference per word instead of one inference per sentence as is the case in \eqref{B2W2V:sent} leading to a drastic drop in computational efficiency.

\subsection{Approach 2 - Paragraphs as context}
Since contextual models are trained on large pieces of texts (generally $\geq 512$ tokens), we instead use paragraphs instead of sentences to obtain the contextual representations.
However, in order to predict target words, we use the contextual embeddings within the sentence only.
Consequently, for this approach, we have
\begin{equation}
  \label{B2W2V:para}
 E_{\text{\textsf{ctx}}}(S,w_t) := \tfrac{1}{|S|}\sum_{w \in S} M(P_S,w),
\end{equation}
where $P_S$ is the paragraph containing sentence~$S$.

In the transfer phase, this approach is more computationally efficient than the previous approach, as
we have to invoke the contextual embedding model $M$ only once for each paragraph as opposed to once for every constituent sentence.
Moreover, it encapsulates the related semantic information in paragraphs in the contextual word embeddings.

We call our models $\xtostatic_{sent}$ in the sentence case \eqref{B2W2V:sent}, and $\xtostatic_{para}$ in the paragraph case  \eqref{B2W2V:para} respectively where \textsc{X} denotes the parent model.

\section{Experiments and Discussion}

\subsection{Corpus Preprocessing and Training}
\label{subsec:preproc}
We use the same English Wikipedia Dump as \citet{Pagliardini2018,Gupta2019BetterWE} to generate distilled \xtostatic representations.
as our corpus for training static word embedding baselines as well as for distilling static word embeddings from pre-trained contextual embedding models. 
We remove all paragraphs with less than~3 sentences or 140 characters, lowercase the characters and tokenize the corpus using the Stanford NLP library \citep{Manning2014TheSC} resulting in a corpus of approximately 54 Million sentences and 1.28 Billion words. 
We then use the Transformers library\footnote{\url{https://huggingface.co/transformers/}} \citep{wolf-etal-2020-transformers} to generate representations from existing transformer models. 
Our \xtostatic representations are distilled from the last representation layers of these models.

We use the same hyperparameter set for training all $\xtostatic$ models, i.e., no hyperparameter tuning is done at all.
We use 12-layer as well as 24-layer pre-trained models using \bert \citep{Devlin2019BERTPO}, \roberta \citep{Liu2019RoBERTaAR} and \gpttwo \citep{Radford2019LanguageMA} architectures as the teacher model to obtain \xtostatic word embeddings. 
All the \xtostatic models use the same set of training parameters except the parent model.
Training hyperparameters are provided in Table~\ref{tab:xtostat_param}.
The distillation/training process employs the lazy version of the Adam optimizer \citep{Kingma2015AdamAM}, suitable for sparse tensors. 
We use a subsampling parameter similar to \fasttext \citep{Bojanowski2016EnrichingWV} in order to subsample frequent target words during training.
Each \xtostatic model was trained using a single V100 32 GB GPU.
Obtaining \xtostatic embeddings from 12-layer contextual embedding models took 15-18 hours while it took 35-38 hours to obtain them from their 24-layer counterparts.

To ensure a fair comparison, we also evaluate \stov, \cbow and \skipgram models that were trained on the same corpus. We do an extensive hyperparameter tuning for these models and choose the one which shows best average performance on the 5 word similarity datasets used in Subsection ~\ref{subsec:unsup}.
These hyperparameter sets can be accessed in Table \ref{tab:baseline_param} where the chosen hyperparameters are shown in bold.
We set the number of dimensions to be 768 to ensure parity between them and the \xtostatic models compared.
We used the \stov library\footnote{\url{https://github.com/epfml/sent2vec}} for training \stov
and the \fasttext library\footnote{\url{https://github.com/facebookresearch/fastText/}} for training \cbow and \skipgram models.
We also evaluate some pre-trained 300 dimensional \glove \citep{Pennington2014GloveGV} and \fasttext \citep{Bojanowski2016EnrichingWV} models in Table~\ref{tab:word_sim}.
The \glove model was trained on Common-Crawl corpus of 840 Billion tokens (approximately 650 times larger than our corpus) while
the \fasttext vectors were trained on a corpus of 16 Billion tokens (approximately 12 times larger than our corpus)). We also extract \ASEblank embeddings from each layer using the same Wikipedia corpus.

We perform two different sets of evaluations. The first set corresponds to  unsupervised word similarity evaluations to gauge the quality of the obtained word embeddings. However, we recognize that there are concerns regarding word-similarity evaluation tasks \citep{Faruqui2016ProblemsWE}
as they are shown to exhibit significant difference in performance when subjected to hyperparameter tuning \citep{Levy2015ImprovingDS}. 
To address these limitations in the evaluation, we also evaluate the \xtostatic embeddings on a standard set of downstream supervised evaluation tasks used in \citet{Pagliardini2018}. 

\begin{table*}[!htb]
\setlength{\tabcolsep}{3pt}%default 6
  \centering
  \resizebox{2.1\columnwidth}{!}{%
\begin{tabular}{l|l|l|c|c|c|c|c|c}
  \toprule
  \midrule
\begin{tabular}{c} Model \textbackslash \\Distilled Model \end{tabular}           & \begin{tabular}{c} Parent Model \textbackslash \\ Other details  \end{tabular}              & Dim. & RG-65      & WS-353 & SL-999 & SV-3500 & RW-2034 & Average \\ \midrule
\midrule
\emph{Existing pre-trained models} & \begin{tabular}{c}  Size  of the \\ training corpus\\ relative to ours \end{tabular}  & \multicolumn{3}{l}{}    \\
\midrule
~\fasttext & 12x & 300 & 0.7669 & 0.596 & 0.416 & 0.3274 & 0.5226 & 0.5276 \\
~\glove & 650x & 300 & 0.6442  & 0.5791 &  0.3764 &  0.2625 &  0.4607 &  0.4646 \\\midrule
\multicolumn{5}{l}{\emph{Models trained by us}}      \\\midrule
~\skipgram     &  N.A.      & 768       & {0.8259}  & 0.7141 & 0.4064 &  0.2722 &  0.4849 &  0.5407 \\
~\cbow         &    N.A.    & 768       & \underline{0.8348}  & 0.4999 & 0.4097 &  0.2626 & 0.4043 & 0.4823 \\
~\stov         &   N.A.     & 768        & 0.7811 &  0.7407 &  0.5034 &  0.3297 &  0.4248 &  0.55594 \\
\midrule
\emph{Models distilled by us} & {Parent Model} & \multicolumn{3}{l}{}     \\\midrule
\midrule
~$\ASEblank$ - best layer per task                               & \bertbase  & 768        & 0.7449(1) & 0.7012(1) & 0.5216(4) & 0.4151(5) & 0.4577(5) & 0.5429(3) \\
~$\ASEblank$ - best overall layer                                 & \bertbase  & 768        & 0.6948(3) & 0.6768(3) & 0.5195(3) & 0.3889(3) & 0.4343(3) & 0.5429(3) \\
\midrule
\rowcolor{LightRed}~$\berttostatic_{sent}$  & \bertbase  & 768        & {0.7421}     & {0.7297}     & \textbf{0.5461}     & \textbf{0.4437}  & \textbf{0.5469} & \textbf{0.6017}  \\
\rowcolor{LightRed}~$\berttostatic_{para}$  & \bertbase  & 768        & {0.7555}     & {\textbf{0.7598}}     & \textbf{0.5384}     & \textbf{0.4317} & \textbf{0.5299} & \textbf{0.6031}\\
\midrule
\midrule
~$\ASEblank$ - best layer per task & \robertabase  & 768        & 0.673(0) & 0.7023(0) & 0.554(5) & 0.4602(4) & 0.5075(3) & 0.5600(0) \\
~$\ASEblank$ - best overall layer & \robertabase  & 768        & 0.673(0) & 0.7023(0) & 0.5167(0) & 0.4424(0) & 0.4657(0) & 0.5600(0) \\
\midrule
\rowcolor{LightRed}~$\robertatostatic_{sent}$ & \robertabase  & 768        & {0.7999}    & \textbf{{0.7452}}     & {{0.5507}}     & \textbf{{0.4658}} & \textbf{0.5496}  &  \textbf{0.6222} \\
\rowcolor{LightRed}~$\robertatostatic_{para}$  & \robertabase  & 768        & {{0.8057}}     & \textbf{\underline{0.7638}}     & \textbf{\underline{0.5544}}     & {\textbf{0.4717}} & \textbf{0.5501} & \textbf{\underline{0.6291}}      \\
\midrule
\midrule
~$\ASEblank$ - best layer per task & \gpttwosmall  & 768   & 0.7013(1) & 0.6879(0) & 0.4972(2) & 0.3905(2) & 0.4556(2) & 0.5365(2)  \\
~$\ASEblank$ - best overall layer  & \gpttwosmall  & 768   & 0.6833(2) & 0.6560(2) & 0.4972(2) & 0.3905(2) & 0.4556(2) & 0.5365(2)  \\
\midrule
\rowcolor{LightRed}~$\gpttwotostatic_{sent}$ & \gpttwosmall  & 768        & 0.7484 & 0.7151 & \textbf{0.5397}  & \textbf{0.4676} & \textbf{\underline{0.5760}} & \textbf{0.6094} \\
\rowcolor{LightRed}~$\gpttwotostatic_{para}$  & \gpttwosmall  & 768        & {{0.7881}}     & 0.7267     & \textbf{0.5417}     & \textbf{\underline{0.4733}} & \textbf{0.5668} & \textbf{0.6193}   \\
\midrule
\bottomrule
\end{tabular}
}
\caption{{\textbf{Comparison of the performance of different embedding methods on word similarity tasks.}
Models are compared using Spearman correlation for word similarity tasks.
All \xtostatic method performances which improve over all $\ASEblank$ methods on their parent model as well as all static models are shown in bold.
Best performance in each task is underlined.
 For all $\ASEblank$ methods, the number in parentheses for each dataset indicates which layer was used for obtaining the static embeddings.
 }}
 \label{tab:word_sim}
\end{table*}

\begin{figure*}[!htb]
\centering
\begin{minipage}[b]{0.75\textwidth}
\includegraphics[scale=0.75]{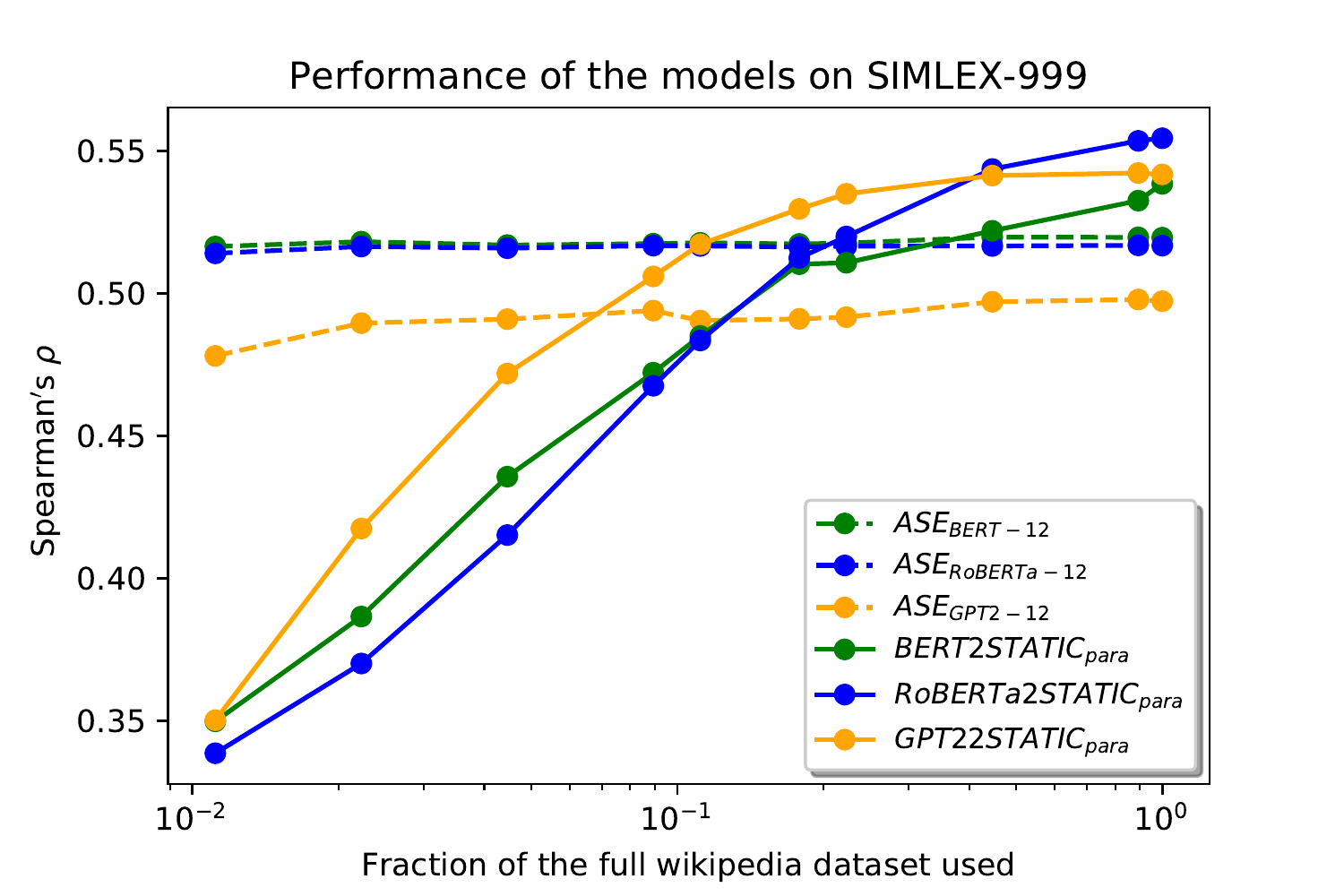}
\end{minipage}\qquad
\begin{minipage}[b]{0.75\textwidth}
\includegraphics[scale=0.75]{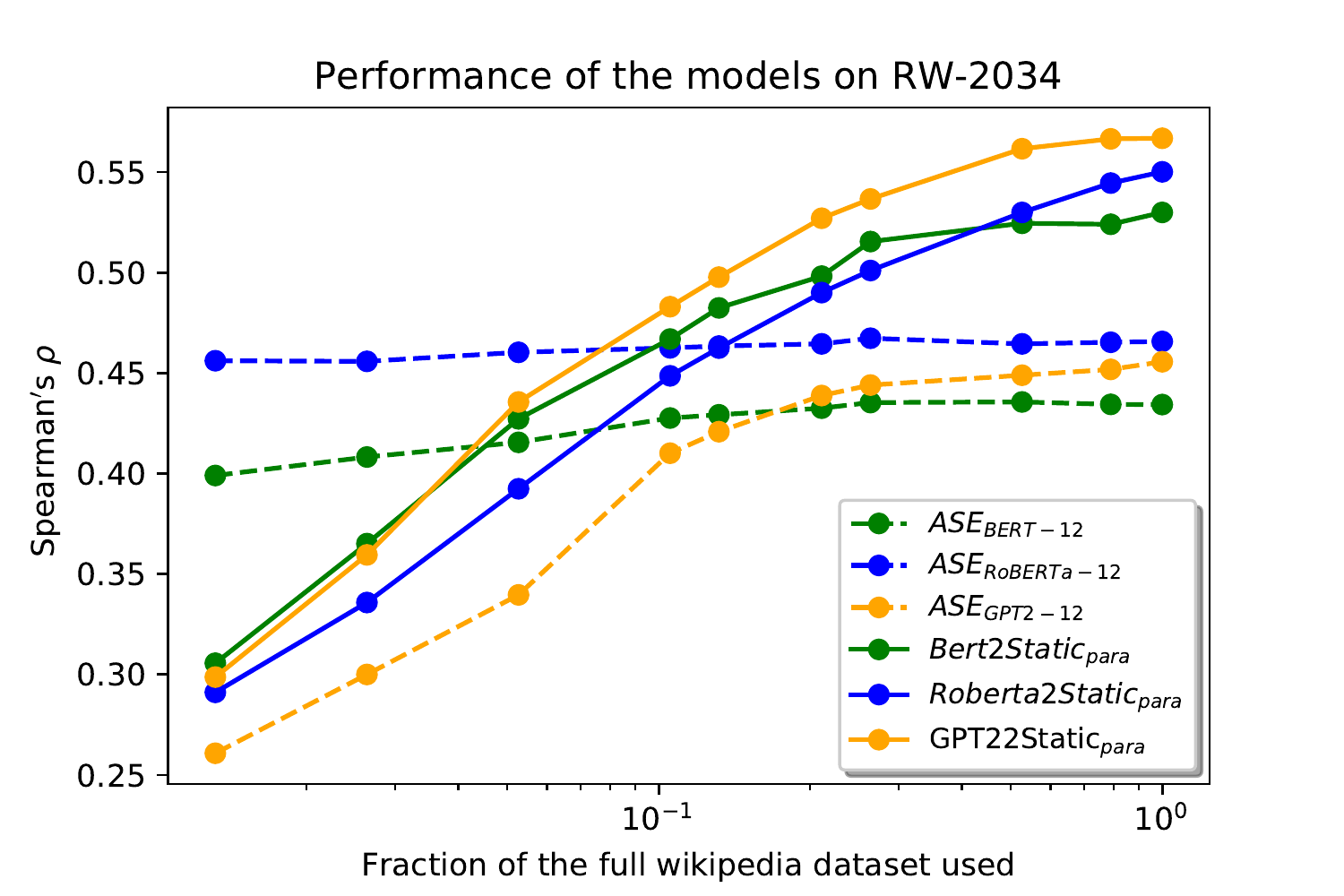}
\end{minipage}
\caption{{\textbf{Effect of corpus size} on the word-embedding quality for \ASEblank best task independent layer and $\xtostatic_{para}$ : In the legend, parent model is indicated in subscript. }}
\vspace{1em}
\label{fig:corpus_size:quality}
\end{figure*}

\begin{table*}[!htb]

\setlength{\tabcolsep}{3pt}%default 6
  \centering
  \resizebox{2.1\columnwidth}{!}{%
\begin{tabular}{l|l|l|l|l|l|l|l|l}
\toprule
\midrule
Embeddings \textbackslash Task & Dim & \multicolumn{1}{c|}{\begin{tabular}[c]{@{}c@{}}CR\\ F1 / Acc.\end{tabular}} & \multicolumn{1}{c|}{\begin{tabular}[c]{@{}c@{}}MR\\ F1 / Acc.\end{tabular}} & \multicolumn{1}{c|}{\begin{tabular}[c]{@{}c@{}}MPQA\\ F1 / Acc.\end{tabular}} & \multicolumn{1}{c|}{\begin{tabular}[c]{@{}c@{}}SUBJ\\ F1 / Acc.\end{tabular}} & \multicolumn{1}{c|}{\begin{tabular}[c]{@{}c@{}}TREC\\ F1 / Acc.\end{tabular}} & \multicolumn{1}{c|}{\begin{tabular}[c]{@{}c@{}}SST-5\\ F1 / Acc.\end{tabular}} & \multicolumn{1}{c}{\begin{tabular}[c]{@{}c@{}}Average\\ F1 / Acc.\end{tabular}} \\
\midrule
\emph{Existing pre-trained models} &  \multicolumn{6}{l}{}    \\
\midrule
\midrule

~\glove                          & 300 & \underline{81.6}/83.2                                                                   & 78.2/78.2                                                                   & 85.1/87.6                                                                     & 90.9/90.9                                                                     & 45.4/86.2                                                                     & 15.5/43.2                                                                      & 66.1/78.1                                                                       \\
~\glove (Twitter)                & 200 & 79.0/80.9                                                                   & 74.1/74.2                                                                   & 82.1/85.0                                                                     & 89.6/89.7                                                                     & 49.1/87.8                                                                     & 13.1/37.5                                                                      & 64.5/75.9                                                                       \\
~\fasttext                       & 300 & 80.3/81.9                                                                   & 78.3/78.4                                                                   & 86.5/88.1                                                                     & 90.9/90.9                                                                     & 45.3/85.9                                                                     & 13.9/43.9                                                                      & 66.2/78.2                                                                       \\
\midrule
\emph{Models trained by us} &  \multicolumn{6}{l}{}    \\
\midrule
~\skipgram                       & 768 & 78.4/80.9                                                                   & 75.2/75.2                                                                   & 83.1/85.8                                                                     & 91.5/91.5                                                                     & 50.2/88.6                                                                     & 13.9/39.0                                                                      & 65.4/76.8                                                                       \\
~\cbow                           & 768 & 75.9/78.5                                                                   & 72.6/72.7                                                                   & 83.3/86.0                                                                     & 85.5/85.5                                                                     & 43.2/85.7                                                                     & 13.4/38.9                                                                      & 62.0/74.6                                                                       \\
~\stov                       & 768 & 79.8/81.2                                                                   & 74.1/74.1                                                                   & 81.0/84.5                                                                     & 89.4/89.4                                                                     & 42.9/84.1                                                                     & 13.2/38.6                                                                      & 63.4/75.3                                                                       \\
\midrule
\emph{Models distilled by us} &  \multicolumn{6}{l}{}    \\
\midrule
\midrule
~$\ASEblank$ - \bertbase(5) & 768 & 81.5/83.0 & 78.5/78.5 & 86.0/86.0 & 91.0/91.0 & 48.3/87.6 & 15.0/42.1 & 66.7/78.0 \\
\rowcolor{LightRed}~$\berttostatic_{sent}$                 & 768 & 80.1/{{82.0}}                                                    & {\textbf{78.9/78.9}}                                                    & \textbf{\underline{87.4}/89.1}                                                            & \textbf{91.8/91.8}                                                            & \textbf{50.6/\underline{88.7}}                                                            & \textbf{16.1}/43.7                                                             & {\textbf{67.5/79.0}}                                                              \\
\rowcolor{LightRed}~$\berttostatic_{para}$                 & 768 & 81.1/\underline{\textbf{83.6}}                                                    & \underline{\textbf{80.8/80.8}}                                                    & \textbf{\underline{87.3}/89.3}                                                            & \textbf{91.6/91.6}                                                            & \textbf{51.8/\underline{89.2}}                                                            & \textbf{16.1/\underline{44.9}}                                                             & \underline{\textbf{68.1/79.9}}                                                              \\
\midrule
\midrule
~$\ASEblank$ - \robertabase(2) & 768 & 78.4/81.2 & 78.3/78.3 & 86.4/88.5 & 89.5/89.5 & 52.0/89.1 & 15.2/43.0 & 66.6/78.3 \\
\rowcolor{LightRed}~$\robertatostatic_{sent}$              & 768 & 76.5/79.6                                                                   & \textbf{80.2/80.2}                                                          & {{85.6/88.0}}                                                      & {\textbf{92.2/92.2}}                                                            & 49.7/\textbf{89.1}                                                            & \textbf{{15.7}}/43.8                                                             & \textbf{66.7/78.8}                                                              \\
\rowcolor{LightRed}~$\robertatostatic_{para}$              & 768 & 80.9/82.3                                                                   & \textbf{80.0/80.1}                                                          & \underline{\textbf{87.3/89.4}}                                                      & \underline{\textbf{92.4/92.4}}                                                            & 49.3/\textbf{88.8}                                                            & \textbf{\underline{16.3}}/43.4                                                             & \textbf{67.7/79.4}                                                              \\
\midrule
\midrule
~$\ASEblank$ - \gpttwosmall(4) & 768 & 81.0/82.1 & 80.1/80.1 & 84.8/86.2 & 91.2/91.2 & 51.0/88.8 &  15.5/42.0 & 67.3/78.4  \\
\rowcolor{LightRed}~$\gpttwotostatic_{sent}$                & 768 & 81.5/\textbf{83.5}                                                                   & \textbf{79.5/79.5}                                                          & \textbf{86.5/88.5}                                                          & \textbf{91.8/91.8}                                                            & \textbf{51.8/89.2}                                                            & \textbf{16.2/43.8}                                                             & \textbf{67.9/79.4} \\
\rowcolor{LightRed}~$\gpttwotostatic_{para}$                & 768 & 81.0/82.6                                                                   & \textbf{79.7/79.7}                                                          & \textbf{86.9/88.8}                                                            & \textbf{92.1/92.1}                                                            & \textbf{\underline{53.0}/89.1}                                                            & \textbf{16.2/44.1}                                                             & \textbf{\underline{68.1}/79.4}\\
\midrule
\multicolumn{1}{c|}{\begin{tabular}[c]{@{}c@{}}\emph{Parent contextual}\\  \emph{models and derivatives} \end{tabular}}& \multicolumn{6}{l}{}     \\
\midrule
\midrule
\bertbase & 768 & 89.6/90.6 & 87.4/87.4 & 89.4/90.8 & 96.7/96.7 & 77.6/94.7 &  30.7/54.0 & 78.6/85.7  \\
\sbertbasenli & 768 & 87.4/88.7 & 83.3/83.3 & 86.8/88.2 & 93.6/93.6 & 41.6/72.2 &  25.3/48.2 & 69.7/79.1  \\
\midrule
\midrule
\robertabase & 768 & 90.0/90.8 & 90.1/90.1 & 89.1/90.6 & 96.3/96.3 & 95.1/99.2 &  34.0/57.6 & 82.4/87.4  \\
\srobertabasenli & 768 & 87.6/88.6 & 86.3/86.3 & 86.8/88.8 & 94.6/94.6 & 52.4/80.6 &  23.7/53.5 & 72.7/82.1  \\
\midrule
\midrule
\gpttwosmall & 768 & 88.5/89.5 & 87.1/87.1 & 87.3/89.1 & 96.1/96.1 & 76.8/94.3 &  30.8/54.5 & 77.8/85.1  \\
\midrule
\bottomrule                                                             
\end{tabular}
}
\caption{{\textbf{Comparison of the performance of different static embeddings on downstream tasks.}
All \xtostatic method performances which improve or are at par over all other static embedding methods and the best $\ASEblank$ layer on their parent model are shown in bold.
Best static embedding performance for each task is underlined. For each ASE method, the number in brackets indicates the layer with best average performance. We use macro-F1 scores and accuracy as the metrics to gauge the performance of models on these downstream tasks. \textbf{Note}: Contextual embeddings for \bertbase, \robertabase and \gpttwosmall in the SOTA section are also fine-tuned while \sbertbasenli and \srobertabasenli are not.
 }}
 \label{tab:downstream}
\vspace{1em}
\end{table*}
\subsection{Unsupervised word similarity evaluation}
\label{subsec:unsup}
To assess the quality of the lexical information contained in the obtained word representations, we use the 4 word-similarity datasets used by \citep{Bommasani2020InterpretingPC}, namely \emph{WordSim353} (353 word-pairs)
\citep{Agirre2009ASO} dataset;  \emph{SimLex-999} (999 word-pairs) \citep{Hill2014SimLex999ES} dataset; \emph{RG-65} (65 pairs) \citep{Joubarne2011ComparisonOS}; and \emph{SimVerb-3500} (3500 pairs) \citep{Gerz2016SimVerb3500AL} dataset as well as
the \emph{Rare Words RW-2034} (2034 pairs) \citep{Luong2013BetterWR} dataset.
To calculate the similarity between two words, we use the cosine similarity between their word embeddings. These similarity scores are compared to the human ratings using Spearman's~$\rho$ \cite{Spearman1904proof} correlation scores. 
We use the tool\footnote{\url{https://github.com/rishibommasani/Contextual2Static}} provided by \citet{Bommasani2020InterpretingPC} to report these results on \ASEblank embeddings. 
It takes around 3 days to obtain \ASEblank representations of the 2005 words in these word-similarity datasets for 12-layer models and around 5 days to obtain them for their 24-layer counterparts on the same machine used for learning \xtostatic representations. 
All other embeddings are evaluated using the MUSE repository evaluation tool\footnote{\url{https://github.com/facebookresearch/MUSE}} \citep{conneau2017word}. 

We perform two sets of experiments concerning the unsupervised evaluation tasks. The first set is the comparison of our \xtostatic models with competing models.
For \ASEblank, we report two sets of results, one which per task reports the best result amongst all the layers and other, which reports the results obtained on the best performing layer on average.

We report our observations in Table~\ref{tab:word_sim}. We provide additional results for larger models in Appendix~\ref{sec:bigmodels}.
We observe that \xtostatic embeddings outperform competing models on most of the tasks.
Moreover, the extent of improvement on SimLex-999 and SimVerb-3500 tasks compared to the previous models strongly highlights the advantage of using improved context representations for training static word representations.

Second, we study the performance of the best \ASEblank embedding layer with respect to the size of corpus used.
\citet{Bommasani2020InterpretingPC} report their results on a corpus size of only up to $N=100,000$ sentences.
In order to measure the full potential of the \ASEblank method, we obtain different sets of \ASEblank embeddings as well as $\xtostatic_{para}$ embeddings from small chunks of the corpus to the full wikipedia corpus itself and compare their performance on SimLex-999 and RW-2034 datasets.
We choose SimLex-999 as it captures true similarity instead of relatedness or association \citep{Hill2014SimLex999ES} and RW-2034 to gauge the robustness of the embedding model on rare words.
We report our observations in Figure \ref{fig:corpus_size:quality}.
We observe that the performance of the \ASEblank embeddings tends to saturate with the increase in the corpus size while $\xtostatic_{para}$ embeddings are either significantly outperforming the \ASEblank embeddings or still show a significantly greater positive growth rate in performance w.r.t. the corpus size.
Thus, the experimental evidence suggests that on larger texts, \xtostatic embeddings will have an even better performance
and hence, \xtostatic is a better alternative than \ASEblank embeddings from any of the layers of the contextual embedding model, and obtains improved static word embeddings from contextual embedding models.

\subsection{Downstream supervised evaluation}
We evaluate the obtained word embeddings on various sentence-level supervised classification tasks. Six different downstream supervised evaluation tasks namely  classification of movie review sentiment(MR) \citep{pang2005seeing},  product  reviews(CR) \citep{hu2004mining}, subjectivity classification(SUBJ) \citep{pang2004sentimental},  opinion polarity (MPQA) \citep{wiebe2005annotating}, question type classification (TREC) \citep{voorhees2002} and fine-grained sentiment analysis (SST-5) \citep{Socher2013RecursiveDM} are employed to gauge the performance of the obtained word embeddings. 

We use a standard CNN based architecture on the top of our embeddings to train our classifier. We use 100 convolutional filters with a kernel size of 3 followed by a ReLU activation function. 
A global max-pooling layer follows the convolution layer. Before feeding the max-pooled output to a classifier, it is passed through a dropout layer with dropout probability of 0.5 to prevent overfitting. 
We use Adam \citep{KingmaB14} to train our classifier. 
To put the performance of these static models into a broader perspective, we also fine-tune linear classifiers on the top of their parent models as well as sentence-transformers \citep{reimers-gurevych-2019-sentence} obtained from \robertabase and \bertbase. 
For the sentence-transformer models, we use the sentence-transformer models obtained by fine-tuning their parent models on the Natural Language Inference(NLI) task using the combination of Stanford NLI \citep{bowman2015large} and the Multi-Genre NLI \citep{williams2018broad} datasets.
The models are refered to as \sbertbasenli and \srobertabasenli in the rest of the paper.

The hyperparameter search space for the fine-tuning process involves the number of epochs (8-16) and the learning rates[1e-4,3e-4,1e-3]. 
Wherever train, validation, and test split is not given, we use 60\% of the data as the training data, 20\% of the data as validation data and the rest as the test data. 
After obtaining the best hyperparameters, we train on the train and validation data together with these hyperparameters and predict the results on the test set.  
For the linear classifiers on the top of parent models, we set the number of epochs and learning rate search space for parent model + linear classifier combination to be [3,4,5,6] and [2e-5,5e-5] respectively. The learning rates in the learning rate search space are lower than those for static embeddings as the contextual embeddings are also fine-tuned and follow the recommendation of \citet{Devlin2019BERTPO}. 
For the sentence-transformer models, we only train the linear classifier and set the number of epochs and learning rate search space to be [3,4,5,6] and [1e-4,3e-4,1e-3] respectively.
We use cross-entropy loss for training all the models.
We use Macro-F1 score and Accuracy to gauge the quality of our predictions. We compare \xtostatic models with all other static models trained from scratch on the same corpus as well as the \glove and \fasttext models used in the previous section. 
We also use existing \glove embeddings trained on tweets(27 billion tokens - 20 times larger than our corpus) \citep{Pennington2014GloveGV} to make the comparison even more extensive. We report our observations in Table~\ref{tab:downstream}. 
For \ASEblank embeddings, we take the layer with best average macro-F1 performance. 

We observe that when measuring the overall performance, with the exception of $\robertatostatic_{sent}$ which has similar average F-1 score to \ASEblank owing to its dismal performance on the CR task, all \xtostatic embeddings outperform their competitors by a significant margin. 
Even though the \glove and \fasttext embeddings were trained on  corpora of one to two magnitudes larger and have a larger vocabulary, their performance lags behind that of the \xtostatic embeddings. 
To ensure statistical soundness, we measure mean and standard deviation of the performance on 6 runs of $\xtostatic_{para}$ model training followed by downstream evaluation along with 6 runs of \ASEblank embedding downstream evaluation with different random seeds in Table~\ref{tab:sixruns} in the Appendix. 
We see that $\xtostatic_{para}$ embeddings outperform \ASEblank by a significant margin.

For both word similarity evaluations and downstream supervised tasks, we observe that $\xtostatic_{para}$ embeddings perform slightly better than $\xtostatic_{sent}$ embeddings. However, since no hyperparameter tuning was performed on the distillation of \xtostatic embeddings, it is hard to discern which \xtostatic variant shows better performance. 
Moreover, owing to the same fact concerning hyperparameter tuning, we expect to see even larger improvements with proper hyperparameter tuning as well as training on larger data. 

\section{Conclusion and Future Work}
This work proposes to augment earlier \wtov-based methods by leveraging recent more expressive deep contextual embedding models
to extract static word embeddings. 
The resulting distilled static embeddings, on an average, outperform their competitors on both unsupervised as well downstream supervised evaluations and thus can be used to replace compute-heavy contextual embedding models (or existing static embedding models) at inference time in many compute-resource-limited applications.
The resulting embeddings can also be used as a task-agnostic tool to measure the lexical information conveyed by contextual embedding models and allow a fair comparison with their static analogues.

Further work can explore extending this distillation framework into cross-lingual domains \citep{schwenk-douze-2017-learning,lample2019cross} as well as using better pooling methods instead of simple averaging for obtaining the context representation, or joint fine-tuning to obtain even stronger static word embeddings.
Another promising avenue is the use of a similar approach to learn sense embeddings from contextual embedding models. 
We would also like to investigate the performance of these embeddings when distilled on a larger corpus along with more extensive hyper-parameter tuning. 
Last but not the least, we would like to release \xtostatic models for different languages for further public use.

\bibliography{bibliography}
\bibliographystyle{acl_natbib}
\clearpage

\appendix

\section{Comparison of multiple downstream runs}
\label{sec:sixruns}
\begin{table}[!htb]

\setlength{\tabcolsep}{3pt}%default 6
  \centering
  \resizebox{1\columnwidth}{!}{%
\begin{tabular}{l|l}
\toprule
\midrule
Embeddings \textbackslash Task & \multicolumn{1}{c}{\begin{tabular}[c]{@{}c@{}}Average\\ Mean F1 / Acc.\end{tabular}} \\
\midrule
~$\ASEblank$ - \bertbase(5) & $67.0 \pm 0.2 / 78.1 \pm 0.2$ \\
\rowcolor{LightRed}~$\berttostatic_{para}$    & ${{68.3 \pm 0.3 /79.9 \pm 0.2}}$ \\
\midrule
~$\ASEblank$ - \robertabase(2) & $67.0 \pm 0.2 /78.2 \pm 0.3$ \\
\rowcolor{LightRed}~$\robertatostatic_{para}$   & ${67.9 \pm 0.2/79.6 \pm 0.3}$    \\
\midrule
\midrule
~$\ASEblank$ - \gpttwosmall(4) & ${67.4 \pm 0.3 / 78.3 \pm 0.3}$  \\
\rowcolor{LightRed}~$\gpttwotostatic_{para}$   & ${68.4 \pm 0.2/80.0 \pm 0.4}$\\
\midrule
                                                      
\end{tabular}
}
\caption{{\textbf{Comparison of the overall performance of $\xtostatic_{para}$ with $\ASEblank$ on downstream tasks.}
Mean and standard deviation of performance on each task over six runs is shown.
 }}
 \label{tab:sixruns}
\vspace{1em}
\end{table}

\section{Experiments on larger models}
\label{sec:bigmodels}
In addition to the smaller 12-layer contextual embedding models, we also obtain \xtostatic word vectors from
larger 24-layer contextual embedding models, once again outperforming their \ASEblank counterparts by a significant margin.
The evaluation results can be accessed in the Table \ref{tab:word_sim_addl}.

\begin{table*}[!htb]
\setlength{\tabcolsep}{3pt}%default 6
  \centering
  \resizebox{2.1\columnwidth}{!}{%
\begin{tabular}{l|l|l|c|c|c|c|c|c}
  \toprule
  \midrule
\begin{tabular}{c} Model \textbackslash \\Distilled Model \end{tabular}           & \begin{tabular}{c} Parent Model \textbackslash \\ Other details  \end{tabular}              & Dim. & RG-65      & WS-353 & SL-999 & SV-3500 & RW-2034 & Average \\ \midrule
\midrule
\textbf{Existing models} & \begin{tabular}{c}  Size  of the \\ training corpus\\ relative to ours \end{tabular}  & \multicolumn{3}{l}{}    \\
\midrule
\fasttext & 12x & 300 & 0.7669 & 0.596 & 0.416 & 0.3274 & 0.5226 & 0.5276 \\
\glove & 650x & 300 & 0.6442  & 0.5791 &  0.3764 &  0.2625 &  0.4607 &  0.4646 \\\midrule
\multicolumn{5}{l}{\textbf{Models trained by us}}      \\\midrule
\skipgram     &  N.A.      & 768       & {0.8259}  & 0.7141 & 0.4064 &  0.2722 &  0.4849 &  0.5407 \\
\cbow         &    N.A.    & 768       & \underline{0.8348}  & 0.4999 & 0.4097 &  0.2626 & 0.4043 & 0.4823 \\
\stov         &   N.A.     & 768        & 0.7811 &  0.7407 &  0.5034 &  0.3297 &  0.4248 &  0.55594 \\
\midrule
\textbf{Models distilled by us} & {Parent Model} & \multicolumn{3}{l}{}     \\\midrule
$\ASEblank$ - best layer per task                               & \bertbase  & 768        & 0.7449(1) & 0.7012(1) & 0.5216(4) & 0.4151(5) & 0.4577(5) & 0.5429(3) \\
$\ASEblank$ - best overall layer                                 & \bertbase  & 768        & 0.6948(3) & 0.6768(3) & 0.5195(3) & 0.3889(3) & 0.4343(3) & 0.5429(3) \\
\midrule
\rowcolor{LightRed}$\berttostatic_{sent}$  & \bertbase  & 768        & {0.7421}     & {0.7297}     & \textbf{0.5461}     & \textbf{0.4437}  & \textbf{0.5469} & \textbf{0.6017}  \\
\rowcolor{LightRed} $\berttostatic_{para}$  & \bertbase  & 768        & {0.7555}     & {\textbf{0.7598}}     & \textbf{0.5384}     & \textbf{0.4317} & \textbf{0.5299} & \textbf{0.6031}\\
\midrule
\midrule
$\ASEblank$  - best layer per task                           & \bertlarge & 1024     &  0.7745(9) & 0.7267(6) & 0.5404(15) & 0.4364(10) & 0.4735(6) & 0.5782(7)\\
$\ASEblank$  - best task independent layer                               & \bertlarge & 1024     &  0.7677(7) & 0.7052(7) & 0.5209(7) & 0.4307(7) & 0.4665(7) & 0.5782(7)\\
\midrule
\rowcolor{LightRed}$\berttostatic_{sent}$  & \bertlarge  & 1024        & {0.8031}     & {{0.7239}}     & \underline{\textbf{0.5675}}     & \textbf{\textbf{0.4692}} & \textbf{0.5595}  & \textbf{0.6247} \\
\rowcolor{LightRed}$\berttostatic_{para}$  & \bertlarge  & 1024        & {{0.8085}}     & \underline{\textbf{0.7652}}     & \textbf{0.5607}     & {\textbf{0.4543}} & \textbf{0.5504}  & \textbf{0.6278}    \\
\midrule
\midrule

$\ASEblank$ - best layer per task & \robertabase  & 768        & 0.673(0) & 0.7023(0) & 0.554(5) & 0.4602(4) & 0.5075(3) & 0.5600(0) \\
$\ASEblank$ - best overall layer & \robertabase  & 768        & 0.673(0) & 0.7023(0) & 0.5167(0) & 0.4424(0) & 0.4657(0) & 0.5600(0) \\
\midrule

\rowcolor{LightRed} $\robertatostatic_{sent}$ & \robertabase  & 768        & {0.7999}    & \textbf{{0.7452}}     & {{0.5507}}     & \textbf{{0.4658}} & \textbf{0.5496}  &  \textbf{0.6222} \\
\rowcolor{LightRed} $\robertatostatic_{para}$  & \robertabase  & 768        & {{0.8057}}     & \textbf{{0.7638}}     & \textbf{{0.5544}}     & {\textbf{0.4717}} & \textbf{0.5501} & \textbf{\underline{0.6291}}      \\
\midrule
\midrule
$\ASEblank$ - best layer per task                          & \robertalarge & 1024    & 0.6782(8) & 0.6736(6) & 0.5526(18) & 0.4571(9) & 0.5385(9) & 0.5680(9)\\
$\ASEblank$ - best task independent layer                               & \robertalarge & 1024    & 0.6738(6) & 0.6270(9) & 0.5437(9) & 0.4571(9) & 0.5385(9) & 0.5680(9)\\
\midrule
\rowcolor{LightRed}$\robertatostatic_{sent}$  & \robertalarge  & 1024        & 0.7677 & 0.7336 & 0.5397 & 0.4576 & \textbf{0.5720}  & \textbf{0.6141}    \\
\rowcolor{LightRed}$\robertatostatic_{para}$  & \robertalarge  & 1024        & 0.7939  &  \textbf{0.7523} & 0.5476 & \textbf{0.4663} & \textbf{0.5739} & \textbf{0.6268}     \\
\midrule
\midrule
$\ASEblank$ - best layer per task & \gpttwosmall  & 768   & 0.7013(1) & 0.6879(0) & 0.4972(2) & 0.3905(2) & 0.4556(2) & 0.5365(2)  \\
$\ASEblank$ - best overall layer  & \gpttwosmall  & 768   & 0.6833(2) & 0.6560(2) & 0.4972(2) & 0.3905(2) & 0.4556(2) & 0.5365(2)  \\
\midrule
\rowcolor{LightRed} $\gpttwotostatic_{sent}$ & \gpttwosmall  & 768        & 0.7484 & 0.7151 & \textbf{0.5397}  & \textbf{0.4676} & \textbf{{0.5760}} & \textbf{0.6094} \\
\rowcolor{LightRed} $\gpttwotostatic_{para}$  & \gpttwosmall  & 768        & {{0.7881}}     & 0.7267     & \textbf{0.5417}     & \textbf{{0.4733}} & \textbf{0.5668} & \textbf{0.6193}   \\
\midrule
\midrule
$\ASEblank$ - best layer per task & \gpttwomedium & 1024  & 0.6574(1) & 0.6957(0) & 0.4988(13) & 0.4226(12) & 0.4566(12) & 0.5155(13) \\
$\ASEblank$ - best task independent layer  & \gpttwomedium & 1024  & 0.5773(13) & 0.6242(13) & 0.4988(13) & 0.4210(13) & 0.4561(13) & 0.5155(13) \\
\midrule
\midrule
\rowcolor{LightRed}$\gpttwotostatic_{sent}$ & \gpttwomedium  & 1024  & 0.7815  & 0.7311 & \textbf{0.5537} & \textbf{0.4774} & \underline{\textbf{0.5939}} & \textbf{0.6275}    \\
\rowcolor{LightRed}$\gpttwotostatic_{para}$  & \gpttwomedium  & 1024  & 0.7907  & 0.7331 & \textbf{0.5488} & \underline{\textbf{0.4850}} & \textbf{0.5828} & \textbf{0.6281}    \\
\midrule
\bottomrule
\end{tabular}
}
\caption{{\textbf{Comparison of the performance of different embedding methods on word similarity tasks.}
Models are compared using Spearman correlation for word similarity tasks.
All \xtostatic method performances which improve over all $\ASEblank$ methods on their parent model as well as all static models are shown in bold.
Best performance in each task is underlined.
 For all $\ASEblank$ methods, the number in parentheses for each dataset indicates which layer was used for obtaining the static embeddings.
 }}
 \label{tab:word_sim_addl}
\end{table*}

\end{document}